\documentclass[letterpaper]{article} 
\usepackage{aaai23}  
\usepackage{times}  
\usepackage{helvet}  
\usepackage{courier}  
\usepackage[hyphens]{url}  
\usepackage{graphicx} 
\usepackage{natbib}  
\usepackage{caption} 
\usepackage{algorithm}
\usepackage{algorithmic}

\usepackage{threeparttable}
\usepackage{graphicx}
\usepackage{subfigure}
\usepackage{multirow}
\usepackage{multicol}
\usepackage{amsmath}
\usepackage{amssymb}
\usepackage{multirow}
\usepackage{diagbox}
\usepackage{wrapfig}
\usepackage{shortbold}
\usepackage{color}

\usepackage{newfloat}
\usepackage{listings}
\DeclareCaptionStyle{ruled}{labelfont=normalfont,labelsep=colon,strut=off} 
\floatstyle{ruled}
\newfloat{listing}{tb}{lst}{}
\floatname{listing}{Listing}
\title{Bayesian Cross-modal Alignment Learning \\ for Few-Shot Out-of-Distribution Generalization}
\title{Bayesian Cross-Modal Alignment Learning \\ for Few-Shot Out-of-Distribution Generalization}
\author {
    Lin Zhu,
    Xinbing Wang,
    Chenghu Zhou,
    Nanyang Ye\thanks{Corresponding author. The full Appendix is available on Arxiv.}
}
\affiliations {
    Shanghai Jiao Tong University, Shanghai, China\\
    zhulin\_sjtu@sjtu.edu.cn, xwang8@sjtu.edu.cn, 
    zhouchsjtu@gmail.com,
    ynylincoln@sjtu.edu.cn
}

\usepackage{bibentry}

\begin{document}

\maketitle

\begin{abstract}
Recent advances in large pre-trained models showed promising results in few-shot learning. However, their generalization ability on two-dimensional Out-of-Distribution (OoD) data, i.e., correlation shift and diversity shift, has not been thoroughly investigated. Researches have shown that even with a significant amount of training data, few methods can achieve better performance than the standard empirical risk minimization method (ERM) in OoD generalization. This few-shot OoD generalization dilemma emerges as a challenging direction in deep neural network generalization research, where the performance suffers from overfitting on few-shot examples and OoD generalization errors. In this paper, leveraging a broader supervision source, we explore a novel Bayesian cross-modal image-text alignment learning method (Bayes-CAL) to address this issue. Specifically, the model is designed as only text representations are fine-tuned via a Bayesian modelling approach with gradient orthogonalization loss and invariant risk minimization (IRM) loss. The Bayesian approach is essentially introduced to avoid overfitting the base classes observed during training and improve generalization to broader unseen classes. The dedicated loss is introduced to achieve better image-text alignment by disentangling the causal and non-casual parts of image features. Numerical experiments demonstrate that Bayes-CAL achieved state-of-the-art OoD generalization performances on two-dimensional distribution shifts. Moreover, compared with CLIP-like models, Bayes-CAL yields more stable generalization performances on unseen classes. 
Our code is available at \url{https://github.com/LinLLLL/BayesCAL}.
\end{abstract}.

\section{Introduction}

Few-shot learning is an emerging research topic that aims to generalize from only a few training samples \cite{wang2020generalizing}.
Despite the success of recent few-shot learning methods on independent and identically distributed (I.I.D) settings \cite{finn2017model, rusu2018meta, sung2018learning, vuorio2019multimodal, fan2021generalized}, these few-shot learning methods would suffer significant performance drop in the presence of domain differences between source and target datasets \cite{chen2019closer, guo2020broader}. This domain shift issue commonly exists in real scenarios, especially in the few-shot setting. For example, it is difficult to construct large training datasets for rare species, there can be huge differences between the training and test environments due to the large randomness of the few-shot setting.

\begin{figure}[t!]
\begin{center}
\centerline{\includegraphics[width=0.47\textwidth]{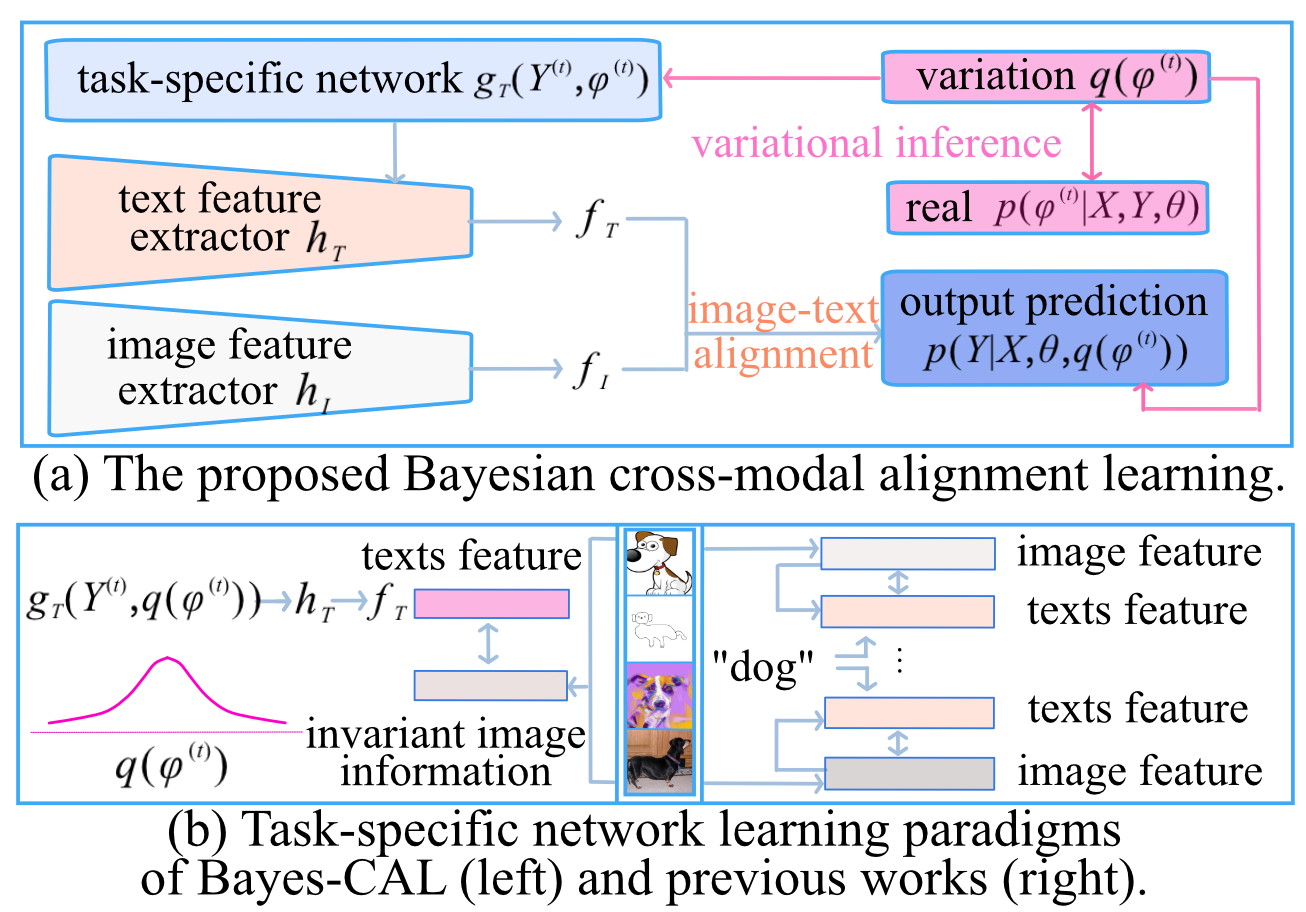}}
\caption{Illustration of the proposed Bayesian cross-modal alignment learning paradigms. 
To avoid overfitting on the few-shot training samples, $\varphi^{(t)}$ is modelled by a variational distribution $q(\varphi^{(t)})$ to approximate its posterior distribution.}
\label{illustration}
\end{center}
\end{figure}

To address this domain shift problem, i.e., training and test data access different conditional distributions, there have been plenty of methods proposed to achieve domain generalization in few-shot learning scenarios\cite{tseng2020cross, liu2021multi, liang2021boosting, zhou2021investigating}. However, most of these studies focused on this issue without considering the different characteristics of distribution shifts between training and test domains. In this paper, we consider a more practical setting of \textit{few-shot Out-of-Distribution (OoD) generalization}, focusing on the few-shot image recognition task of generalizing under two major types of distribution shifts (as described in OoD-Bench \cite{OoDBench}). 
Specifically, models usually have access to multi-domain "K-shot N-way" training samples of OoD data dominated by diversity shift or correlation shift. According to \cite{OoDBench}, the diversity shift is defined by the support set's differences on latent environment's distributions (e.g., changes in the image style), and the correlation shift is defined by the probability density functions' differences caused by spurious correlations.
Under this condition, models are expected to learn feature distributions only from the seen domains and generalize to unseen domains within the same downstream task.

In recent few-shot learning methods, studies\cite{rahman2020improved, zhou2022learning, zhang2021vinvl, zhu2021semantic} based on large-scale vision-language pre-trained models 
achieved striking performances in various downstream tasks. 
The conventional visual deep learning models that only focus on closed-set visual concepts, are susceptible to overfitting on the pre-defined list of classes due to their limited supervision source \cite{zhou2022cocoop}. In contrast, vision-language models leverage a broader source of supervision coming from natural language, which has been proven effective in learning transferable representations \cite{jia2021scaling, radford2021learning}.

A widely utilized method in vision-language pre-trained models is the cross-modal representation learning \cite{fang2022learning, li2020oscar, wehrmann2020adaptive, zheng2020background}. 
Inspired by the advantages of the cross-modal alignment and Bayesian methods \cite{lin2022bayesian} for alleviating overfitting, instead of focusing on improving learning algorithms for learning OoD generalizable image features,
we propose the Bayesian cross-modal alignment learning method (Bayes-CAL) to achieve few-shot OoD generalization.
Since fine-tuning the entire model is impractical and might damage the well-learned representation space \cite{zhou2022cocoop} and adjusting the text representations are more flexible, as shown in Figure~\ref{illustration} (a), we design the model architecture as only a task-specific network in text feature extractor is tuned in each specific downstream task. 
In this paper, we instantiate the text representation learning of Bayes-CAL in three methods--prompt leaning \cite{ding2021prompt}, directly utilizing learnable vectors, and Word2Vector \cite{mikolov2013distributed} to detail how Bayes-CAL works on few-shot OoD generalization. 
Our key contributions are as follows:
\begin{enumerate}
    \item 
    We propose a Bayesian treatment for cross-modal alignment learning for few-shot OoD generalization. The superiority of the Bayesian method is demonstrated by stable generalization performances on unseen classes. We have also carefully designed experiments to gain insight into the superiority of image-text alignment learning.
    
    \item Under the proposed architecture (see Figure~\ref{pipeline}),
    gradient orthogonalization loss is introduced to achieve better alignment learning by disentangling image features. Invariant risk minimization (IRM) loss is utilized to improve OoD generalization ability further.
    
    \item Bayes-CAL has achieved state-of-the-art performances on OoD-Bench datasets with both diversity shift and correlation shift, especially 10\%-20\% performance improvements compared with algorithms in OoD-Bench \cite{OoDBench}. Moreover, it outperforms CLIP-like solid models by more stable generalization performance on both I.I.D and OoD unseen classes.
    
\end{enumerate}

\section{Related Work}
\subsection{Foundation Models}
In this paper, we focus on foundation models of large-scale vision-language pre-training, which have recently emerged \cite{gu2021ppt, dai2021coatnet, radford2021learning} for various image-text retrieval tasks.
Especially for image recognition, a contrastive language image pre-training model (CLIP, \cite{radford2021learning}) is proposed. In CLIP, images and texts are encoded to the feature space. Then, the model is optimized to maximize the similarity of image features and texts features. There are also many efficient CLIP-based models to enhance generalization performance via prompt tuning or image feature adapters. Prompt tuning is a type of method to get better vision-language alignment via only fine-tuning the input prompts, such as CoOp \cite{zhou2022learning}, CoCoOp \cite{zhou2022cocoop}, DPLCLIP \cite{APCLIP}, etc. An alternative path is to conduct fine-tuning with image feature adapters on visual feature space, like CLIP-Adapter \cite{gao2021clip} and Tip-Adapter \cite{zhang2021tip}. In this paper, we mainly focus on the methods that fine-tune on the semantic space of texts. For example, 
with only few-shot samples for learning, CoOp improved significantly in generalization ability over intensively-tuned manual prompts via prompt learning. However, a critical problem of CoOp is identified as its learned context is not generalizable to unseen classes \cite{zhou2022cocoop}.
Motivated by learning generalization prompts, CoCoOp is proposed to achieve generalization on unseen classes via conditional prompt learning. Nevertheless, 
CoCoOp requires for each image an independent forward pass of instance-specific prompts, which significantly decreases its training efficiency. 
Another method Domain Prompt Learning (DPLCLIP), is proposed to guide CLIP to domain transfer learning by domain inference in the form of prompt generating. It captures domain shifts by extracting information from image features, which limits its generalization ability in distribution shifts that can not be extracted efficiently from few-shot images. 
For more information on foundation models, we refer readers to this survey \cite{du2022survey}. 

\subsection{Out-of-Distribution Generalization Algorithms}

Plenty of methods for OoD generalization have been proposed recently. Typically, they can be categorized into three types---1) Invariant learning-based methods, such as invariant risk minimization (IRM, \cite{arjovsky2019invariant}), invariant risk minimization games (IRM-Games \cite{ahuja2020invariant}), etc. 
2) Domain generalization methods, such as the Jigsaw method (Jigsaw, \cite{carlucci2019domain}), the representation self-challenging method (RSC, \cite{huang2020self}), etc. 
For more detailed information, we refer readers to this survey \cite{wang2022generalizing}. 
3) Stable learning methods, such as the sample reweighting method \cite{shen2020stable}. 
Although these methods yield OoD generalization performance improvement to some extent, it has been recently demonstrated that they are pretty hard to systematically beat the standard ERM method \cite{gulrajani2021in, OoDBench}.
Furthermore, learning from the few-shot OoD samples is much more challenging due to their large randomness, and the few-shot OoD generalization under two major distribution shifts is rarely understood.

\section{Methodology}
\begin{figure*}[!t]
\vskip 0.2in
\begin{center}
\centerline{\includegraphics[width=1.0\textwidth]{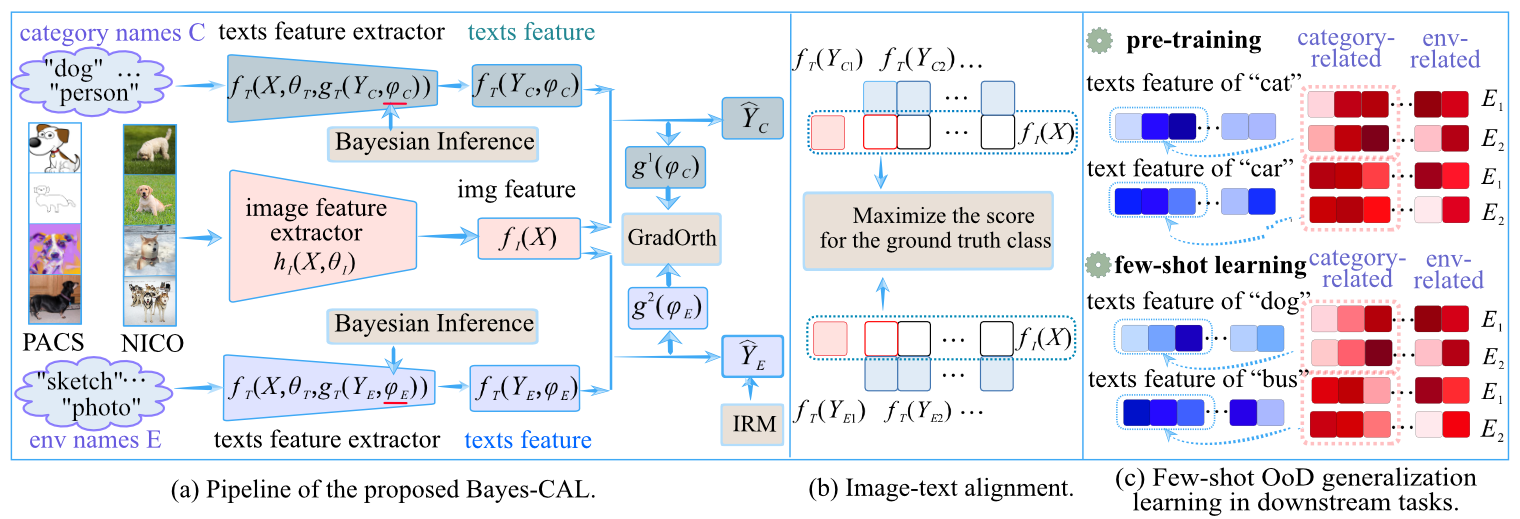}}
\caption{
(a) Pipeline of the proposed Bayes-CAL. The top and the bottom branches are used to generate category-related texts feature and environment-related texts feature, respectively. 
(b) The image-text alignment process. The cross-modal alignment is achieved by maximizing the cosine similarities between the image feature $f_I(X)$ and its corresponding texts feature $f_T(Y_{gt})$ (where $Y_{gt}$ is the ground-true class name of $X$).
(c) Few-shot OoD generalization learning in downstream tasks. 
The embeddings of new classes "dog" and "bus" are very close to pre-training classes "cat" and "car" in the natural language models, respectively.
While in the few-shot OoD setting, the image feature of the same category extracted by different environments may be various. This results in previous models hard to achieve image-text alignment since conditional information (especially for data dominated by complex correlation shift) can not be efficiently extracted from the image features.
Benefiting from the image-text alignment incorporating the proposed loss, regardless of various environmental information, texts features can be aligned with the disentangled image features by a few tuning steps of texts representations. 
Note that the element positions of the category-related and environment-related features can be random.
}
\label{pipeline}
\end{center}
\end{figure*}
We propose a novel Bayesian cross-modal alignment learning method (Bayes-CAL) for few-shot OoD generalization. 
Unlike CoCoOp and DPLCLIP that fine-tune task-specific parameters by incorporating the conditional information extracted from image features, we fine-tune on the semantic space by enforcing domain-invariant alignment under the proposed regularization terms. Moreover, the Bayesian treatment is specially introduced to substantially alleviate overfitting. Based on the domain-invariant information disentangled from the image features, the distributions of the task-specific parameters are estimated.
Without a query of a large amount of GPU memory like CoCoOp in every run, the proposed Bayes-CAL is simple yet efficient, making fine-tuning on few-shot samples practical in the few-shot OoD setting. 
An overview of the Bayes-CAL is shown in Figure~\ref{pipeline}.




\subsection{Preliminary on Few-Shot Cross-Modal Alignment Learning}

In a specific downstream classification task $t$, models are supposed to learn from the few-shot OoD training data $D^{(t)}=\left\{X^{(t)}, Y^{(t)}\right\}$ ($X^{(t)}$ denotes the input image and $Y^{(t)}$ denotes the category label) and generalize well on new domains within the same task $t$.

As shown in Figure~\ref{illustration} (a), incorporated with task-specific network $g_T(Y^{(t)}, \varphi^{(t)})$, the text feature extractor $h_T(\cdot)$ parameterized with $\theta_T$ is utilized to extract texts features, the image feature extractor $h_I(\cdot)$ parameterized with $\theta_I$ is used to extract image features. As discussed in the Introduction, the parameters $\theta = (\theta_I, \theta_T)$ are shared across all tasks. 
Given the downstream classification task $t$, with both $h_T(\cdot)$ and $h_I(\cdot)$ fixed, the task-specific parameter $\varphi^{(t)}$ is optimized to achieve few-shot OoD generalization. 

In contrast with previous works that fine-tune text representations by conditional prompt learning, our method does not require additionally interacting with image features. As shown in Figure~\ref{illustration} (b), the conditional prompt learning paradigm generates input-specific or domain-specific prompts by learning from both the label name "dog" and the corresponding image feature. However, this paradigm would suffer a significant performance drop when conditional information cannot be efficiently extracted (especially for data dominated by complex correlation shifts).

In this paper, instead of generating conditional prompts, 
we fine-tune text representations based on the domain-invariant image information to achieve better alignment. To further avoid overfitting on the pre-defined class observed during training, we model $\varphi^{(t)}$ by a variational distribution $q(\varphi^{(t)})$ to approximate its posterior distribution $p(\varphi^{(t)}|X^{(t)}, Y^{(t)},\theta)$. Given the learned invariant image information, Bayes-CAL estimates the distribution of $\varphi^{(t)}$ across domains, which incorporates richer semantic information compared with the determined values adopted in previous works. 
Hence, the probability distribution of a sample $(x_n^{(t)}, y_n^{(t)})$ can be represented as:
\begin{equation}
\begin{aligned}
p(y_n^{(t)}|x_n^{(t)}, \theta)
&\propto p(y_n^{(t)}|x_n^{(t)}, \theta, g_T(y_n^{(t)}, q(\varphi^{(t)})))\\
&p(g_T(y_n^{(t)}, q(\varphi^{(t)}))|\theta, D^{(t)})
\end{aligned}
\label{eq:alignment_few_shot_learning_paradigm}
\end{equation}

\subsection{Bayesian Cross-Modal Alignment Learning}
In the following contents, we omit $t$ from the corresponding  mathematical expressions for convenience.

As illustrated in Figure~\ref{pipeline} (a), to disentangle the causal and non-casual parts from the image feature $\boldsymbol{f_I}$,
the category-related text branch $h_T(X, \theta_T, g_T(Y, \varphi_C))$  and the environment-related text branch $h_T(X, \theta_T, g_T(Y, \varphi_E))$ are introduced to learn category-related information and environment-related information, respectively.
Then the similarity between the texts feature (category-related texts feature $f_T(Y_{C}, \varphi_C)$ or environment-related texts feature $f_T(Y_{E}, \varphi_E)$) and the image feature $\boldsymbol{f_I}$ are measured to calculate the final output of category prediction probability $\boldsymbol{\hat{Y}_{C}}$ and environment prediction probability $\boldsymbol{\hat{Y}_{E}}$. 

Figure~\ref{pipeline} (b) illustrates a showcase of the image-text alignment process. In this paper, we compute the cosine similarity for each image-text alignment and then input them into classical cross-entropy loss. And thus, the cross-modal alignment is achieved by maximizing the cosine similarity of the image feature and the ground-true texts feature.

Figure~\ref{pipeline} (c) shows the model's working mechanism in the few-shot OoD generalization learning. In the pre-training process, alignment learning usually makes visually similar classes semantically embedded more closely (see Appendix C). 
The image feature extracted by $h_I$ may contains category-related information $\boldsymbol{f_{I_C}}$ as well as environment-related information $\boldsymbol{f_{I_E}}$. The $\boldsymbol{f_{I_C}}$ is invariant across domains, but $\boldsymbol{f_{I_E}}$ may be various, resulting in previous models hard to achieve efficient image-text alignment.

\paragraph{Gradient Orthogonal Loss}
To disentangle the category-related and context-related information from the image feature, we introduce an orthogonalization regularization, i.e., the two gradients (w.r.t. the image feature $\boldsymbol{f_I}$) of losses for predicting category and environment labels should be orthogonal.

As shown in Figure~\ref{pipeline} (b), 
the cosine similarities between the image features $\boldsymbol{f_I}$ and the category-related texts features $\boldsymbol{f_{T_C}}$ generated by $h_T(X, \theta_T, g_T(Y_C, \varphi_C))$ are calculated to obtain the category prediction probability $\boldsymbol{\hat{Y}_{C}}$. We perform similar operations to get the environment prediction probability $\boldsymbol{\hat{Y}_{E}}$.
And then, the image feature $\boldsymbol{f_I}$ can be disentangled into the category-related and environment-related parts by orthogonalizing the two gradients of the cross-entropy losses $\ell(\boldsymbol{\hat{Y}_{C}}, Y_{C})$ and $\ell(\boldsymbol{\hat{Y}_{E}}, Y_{E})$ with respect to $\boldsymbol{f_I}$. Therefore, the direction that changes the category loss most quickly will not change the environment loss from  $\boldsymbol{f_I}$ and vice versa. In other words, the directions in which the gradients of the two losses change fastest are not on the same hyperplane, which enforces the two branches to pay attention to different parts of the image feature $\boldsymbol{f_I}$ if it contains both category and environment information.
Specifically, let $\mathcal{G}^{1}(\varphi_{C})=\nabla_{\boldsymbol{f_I}}\ell(\boldsymbol{\hat{Y}_{C}}, Y_{C})$ and $\mathcal{G}^{2}(\varphi_E)=\nabla_{\boldsymbol{f_I}}\ell(\boldsymbol{\hat{Y}_{E}}, Y_{E})$ be the gradients of the category prediction loss and the environment prediction loss with respect to $\boldsymbol{f_I}$, respectively. Thus, the following gradient orthogonal regularization term is introduced:
\begin{equation}
\mathcal{L}_{\text {orth}}\left(\varphi_{C}, \varphi_{E} \right)=\left(\frac{\mathcal{G}^{1}\left(\varphi_{C}\right)}{\left\|\mathcal{G}^{1}\left(\varphi_{C}\right)\right\|} \cdot \frac{\mathcal{G}^{2}\left(\varphi_{E}\right)}{\left\|\mathcal{G}^{2}\left(\varphi_{E}\right)\right\|}\right)^{2}
\end{equation}

\paragraph{Invariant Risk Minimization}
Invariant risk minimization is proposed to achieve invariant predictions across different environments \cite{ahuja2020invariant}. It aims to learn a stable data representation $\varphi:\mathcal{X}\rightarrow\mathcal{H}$ and an invariant predictor $\omega: \mathcal{H}\rightarrow\mathcal{Y}$ across for all training distributions $\mathcal{E}_{\mathrm{tr}}$. Mathematically, it can be written as:
\begin{equation}
\begin{aligned}
&\min _{\varphi: \mathcal{X} \rightarrow \mathcal{H} \atop w: \mathcal{H} \rightarrow \mathcal{Y}} \sum_{e \in \mathcal{E}_{\mathrm{tr}}} \ell^{e}(w \circ \varphi)\\
&\text { subject to } \quad w \in \underset{\bar{w}: \mathcal{H} \rightarrow \mathcal{Y}}{\arg \min } \ell^{e}(\bar{w} \circ \varphi), \text { for all } e \in \mathcal{E}_{\mathrm{tr}}
\end{aligned}
\label{eq:ERM}
\end{equation}
where $\ell^e$ is the ERM risk of the training environment $e$. Since each constraint calls an inner optimization routine in Equation~(\ref{eq:ERM}), the author proposes to instantiate the challenging bi-leveled optimization problem by adding a gradient norm penalty $\left\|\nabla_{w \mid w=1.0} \ell^{e}(w \cdot \varphi)\right\|^{2}$ as regularization term. So, the optimization problem is changed into the following form:
\begin{equation}
\begin{aligned}
\min _{\varphi: \mathcal{X} \rightarrow \mathcal{Y}} \sum_{e \in \mathcal{E}_{\mathrm{tr}}} \ell^{e}(\varphi)+\lambda \cdot\left\|\nabla_{w \mid w=1.0} \ell^{e}(w \cdot \varphi)\right\|^{2}
\end{aligned}
\label{eq}
\end{equation}

In Bayes-CAL, we utilize IRM to train the task-specific network $g_T(Y_C, \varphi_C)$ suitable for all training distributions to learn domain-invariant category-related text representations. Given the
category prediction probability $\boldsymbol{\hat{Y}_C}$,
the IRM regularization term is defined as:
\begin{equation}
\mathcal{L}_{\text {IRM}}(\varphi_{C}) = \sum_{e\in \mathcal{E}_{\mathrm{tr}}} \|\nabla_{w|w=1} \ell_{e}(w*\boldsymbol{\hat{Y}_{C}}, Y_{C})\|^2
\label{eq}
\end{equation}

\paragraph{Bayesian Learning}
To further enhance the generalization ability of the proposed method, Bayesian inference is utilized to
overcome overfitting on the pre-defined clasess observed during training. 
It is known that variational inference is appealing when dealing with overfitting in few-shot learning since applying Monte Carlo sampling to the Bayesian neural network can be computationally infeasible. From Bayesian perspective, these unknown parameters $\varphi_{C}$ and $\varphi_E$ can be viewed as latent variables that follow some prior distribution $p(\varphi_{C})$ and $p(\varphi_{E})$, respectively. In order to infer $\varphi_{C}$ and $\varphi_{E}$, training samples containing the information about the unknown parameters are utilized to approximate the posterior distribution $p(\varphi_{C}, \varphi_{E}|X, Y_{C}, Y_{E})$.

In this paper, under the mean-field assumption partitioning the variables into independent parts, we have,
\begin{equation}
p(\varphi_{C}, \varphi_{E})=\prod_{i=1}^{K}\prod_{j=1}^{K} p_{i}\left(\varphi_{C_{i}}\right)p_{j}\left(\varphi_{E_{j}}\right)
\end{equation}
where $K$ is the total number of learnable parameters $\varphi_C$ or $\varphi_E$. And we assume each $\varphi_{C_i}$ follows a Gaussian distribution $\mathcal{N}(\mu_{i1}, \sigma_{i1})$ with the posterior distribution of $\mathcal{N}(\mu_{i2}, \sigma_{i2})$. For each $\varphi_{E_{i}}$, we have the same assumption.

Variational inference seeks for a variational distribution $q(\varphi_{C}, \varphi_{E})$ to approximate $p(\varphi_{C}, \varphi_{E}|X, Y_{C}, Y_{E})$ by minimizing the Kullback-Leibler divergence $\mathbb{D}_{\mathrm{KL}}[q(\varphi_{C}, \varphi_{E}) \| p(\varphi_{C}, \varphi_{E}|X, Y_{C}, Y_{E})]$ between them. This can be equivalent to minimizing the negative value of the evidence lower bound (ELBO):
\begin{equation}
\begin{aligned}
&\mathcal{L}_{\text{Bayes}}
=-\mathbb{E}_{q(\varphi_{C}, \varphi_{E})}\left\{[\log p(Y_{C}, Y_{E} \mid X,  \varphi_{C}, \varphi_{E})]  \right.\\
&\left.+\mathbb{D}_{\mathrm{KL}}[q(\varphi_{C}, \varphi_{E}) \| p(\varphi_{C}, \varphi_{E}\mid X,  Y_{C}, Y_{E})]\right\}
\end{aligned}
\label{L}
\end{equation}
The first term $\mathcal{L}_{1}$ in Equation~(\ref{L}) is the expectation of negative log-likelihood. According to Monte Carlo sampling and adding the regularization terms to the negative log-likelihood, we approximate $\mathcal{L}_{1}$ by:
\begin{equation}
\begin{aligned}
\mathcal{L}_{1} &\approx \frac{1}{N}\sum_{i}^{N}(\ell\left(\boldsymbol{\hat{Y}_{C}},Y_{C}\right) + \lambda_1 \ell\left(\boldsymbol{\hat{Y}_{C}},Y_{E}\right) \\
&+ \lambda_2 \mathcal{L}_{\text {IRM}}(\varphi_{C}) + \lambda_3 \mathcal{L}_{\text {orth}}\left(\varphi_{C}, \varphi_{E} \right))
\end{aligned}
\label{L1}
\end{equation}
According to the Mean-Field assumption, the second term $\mathcal{L}_{2}$ in Equation~(\ref{L}) can be calculated as:
\begin{equation}
\begin{aligned}
\mathcal{L}_{2} =\mathbb{D}_{\mathrm{KL}}(\varphi_{C}) + \mathbb{D}_{\mathrm{KL}}(\varphi_{E})
\end{aligned}
\label{L2}
\end{equation}
where based on the Gaussian distribution assumption, $\mathbb{D}_{\mathrm{KL}}(\theta)$ is simplified as:
\begin{equation}
\begin{aligned}
\mathbb{D}_{\mathrm{KL}}(\varphi_{C})=\sum_{i=1}^{K} \log \frac{\sigma_{i1}}{\sigma_{i 2}}+\frac{1}{2}\left(\sigma_{i2}^{2}+\left(\mu_{i2}-\mu_{i1}\right)^{2}\right) / \sigma_{i1}^{2}
\end{aligned}
\end{equation}
and $\mathbb{D}_{\mathrm{KL}}(\varphi_E)$ can be computed similarly.

By integrating Equation~(\ref{L1}) and Equation~(\ref{L2}) into Equation~(\ref{L}), we obtain the final learning objective as follows:
\begin{equation}
\label{objective}
\begin{aligned}
\mathcal{L}_{\text{Bayes}}
&=\frac{1}{N}\sum_{i}^{N}(\ell\left(\hat{Y}_{C},Y_{C}\right) + \lambda_1 \ell\left(\hat{Y}_{E},Y_{E}\right) \\
&+ \lambda_2 \mathcal{L}_{\text {IRM}}(\varphi_{E}) + \lambda_3 \mathcal{L}_{\text {orth}}\left(\varphi_{C}, \varphi_{E}\right)) \\
&+ \mathbb{D}_{\mathrm{KL}}(\varphi_{C}) + \mathbb{D}_{\mathrm{KL}}(\varphi_{E})
\end{aligned}
\end{equation}

\section{Experiment Results}
We evaluate Bayes-CAL in the following three settings:
1) First, without loss of generality, we instantiate Bayes-CAL by the prompt learning method of CLIP as a showcase, and compare it with OoD generalization algorithms in OoD-Bench \cite{OoDBench} and several powerful CLIP-like models. 2) Then, we do a series of ablation studies and evaluate its base-to-new generalization ability. 3) Furthermore, to validate its working mechanism for few-shot OoD generalization, we instantiate the text branches of Bayes-CAL with other methods. Despite the strong Transformer-based pre-trained models of the two text branches, we substitute them directly with learnable vectors (LV for short) or word embeddings from Word2Vector (W2V for short). 
We conduct fair comparisons with the conventional visual deep network on convergence speed and OoD generalization performances. 

\subsubsection{Datasets}
We evaluate Bayes-CAL on datasets that cover both diversity shift and correlation shift: datasets dominated by correlation shift (NICO \cite{he2021towards} and ColoredCatsDogs),
and datasets dominated by diversity shift (PACS \cite{li2017deeper} and VLCS \cite{torralba2011unbiased}). ColoredCatsDogs (CCD for short) has spurious correlations with the background color (green or red), constructed in a similar principle as ColoredMNIST \cite{arjovsky2019invariant} but with images of cats and dogs disturbed by Gaussian noise to increase complexity.

Following CLIP, we train models on a few-shot training data and evaluate the original test set. Due to the difficulty of the task and the large randomness of OoD data, for each category, 
we randomly sample a 16-shot training set and a 16-shot validation set (an 8-shot training set and a 64-shot validation set for NICO due to the validation accuracy up to 100\% if the validation set is too small) from each domain. Note that the 16-shot training set is equally sampled from every domain. 
For all experiments, we set the max epoch as 30 unless otherwise specified and conduct three independent experiments with random seeds (1, 2, 3) to exclude the effects of randomness.

\subsubsection{Experiment Protocol}
Implementing experiments based on CoOp's code, we first give Bayes-CAL's hyper-parameter setting introduced from CoOp. Throughout the experiments, the ResNet-50 model \cite{he2016deep} is used as the vision backbone. The number of context tokens
is set as 16, the class token position (CTP) is a hyper-parameter that can be set as "end" or "middle", and the class-specific context (CSC) can be set as "True" or "False".
The three additional hyper-parameters introduced by our paradigm are $\lambda_1$, $\lambda_2$, and $\lambda_3$, corresponding to the coefficients of the three regularization terms. For fair comparisons, a similar model evaluation protocol as OoD-Bench is used---a 20-times random search for each of 3 pairs of weight initialization and training-validation data. Finally, we report the mean and standard error values on the original test set.

\subsubsection{Competitors}
We compare Bayes-CAL with the \textbf{top three} algorithms for each dataset in OoD-Bench on the four datasets.
Since the proposed method is instantiated by the large pre-trained model of CLIP, we mainly evaluate the OoD generalization performance of CLIP-like alignment learning methods based on text representation fine-tuning, i.e., CLIP, CoOp, CoCoOp, and DPLCLIP.
The performances of the three CLIP-based methods are evaluated under the same experimental setting. Note that the OoD-Bench results on CCD and VLCS are reproduced by ourselves.


\subsection{OoD Generalization on 2D Distribution Shifts}

\begin{table}
\centering
\resizebox{\linewidth}{!}{
\begin{tabular}{llll} 
\hline
Algorithm                                                  & NICO              & Algorithm                                               & CCD                \\ 
\hline
\textbf{Bayes-CAL}                                                & \textbf{98.33(0.6)} & \textbf{Bayes-CAL}                                            & \textbf{69.00(0.8)}  \\
CLIP                                                       & 96.75(0.0)          & CLIP                                                    & 65.00(0.0)           \\
CoOp                                                       & 95.92(0.7)          & CoOp                                                  & 55.21(10.6)           \\
 CoCoOp                                                          & 98.00(0.8)   & CoCoOp                                                 & 56.25(6.8)      \\
DPLCLIP       &97.42(1.4)
&DPLCLIP  &57.29(10.3)
\\

\hline
ANDMask    & 72.20(1.2)          & IRM        & 51.72(0.5)           \\
GroupDRO & 71.83(0.8)          & ERDG      & 51.71(2.0)           \\
ERM          & 71.44(1.3)          & SagNet          & 44.41(0.0)           \\

\hline
\end{tabular}
}
\caption{Performances on NICO and CCD.}
\label{cor}
\end{table}

\subsubsection{Experiment on the Correlation Shift Datasets}
NICO and CCD are typical datasets with correlation shift. 
Following the same OoD validation for NICO and test-domain validation for CCD as in OoD-Bench.
As shown in Table~\ref{cor}, Bayes-ACL obtains far higher accuracy compared with the methods in OoD-Bench, with more than 20\% improvements for NICO and more than 15\% improvements for CCD with statistical significance. This demonstrates the superiority of Bayes-CAL incorporating the large-scale pre-trained
vision-language model. 
Moreover, Bayes-CAL outperforms CLIP-like models and obtains far better results on CCD with strong spurious correlations. 
As shown in Figure~\ref{landscape} (a), the test accuracy on CCD of Bayes-CAL is stable as the training epoch increases. The result of CoOp, however, shows an obvious trend of overfitting on training data, since its test accuracy gradually decreases during the training process.


\subsubsection{Experiment on the Diversity Shift Datasets}

\begin{table}
\centering
\resizebox{\linewidth}{!}{
\begin{tabular}{llll} 
\hline
Algorithm                                                & PACS              & Algorithm                                                   & VLCS               \\ 
\hline
\textbf{Bayes-CAL}                                              & \textbf{91.82(0.9)}         & \textbf{Bayes-CAL}                                                  & \textbf{78.06(1.5)}  \\
CLIP                                                    & 90.76(0.0)          & CLIP                                                      & 75.07(0.0)           \\
CoOp                                                  & 91.47(0.6) & CoOp                                                        & 72.79(5.4)           \\ 
CoCoOp                                                    & 91.81(0.6)  & CoCoOp                                                & 76.25(1.0)      \\
DPLCLIP      &89.58(0.7)
&DPLCLIP  &74.25(6.1)
\\

\hline
RSC             & 82.82(0.4)          & RSC                  & 77.14(0.5)           \\
VREx & 81.78(0.1)          & ANDMask     & 77.13(0.4)           \\
MMD           & 81.72(0.2)          & MLDG              & 76.25(0.6)            \\

\hline
\end{tabular}
}
\caption{Performances on PACS and VLCS.}
\label{div}
\end{table}
PACS and VLCS are two common domain generalization benchmark datasets with diversity shift as shown by OoD-Bench. We follow the same training-domain validation experimental protocol as in OoD-Bench.
The mean accuracy of the four tests on different domains is shown in Table~\ref{div}. It is shown that our method outperforms all models in OoD-Bench on PACS and VLCS, especially with an improvement of around
10\% on PACS. Bayes-CAL gets significant improvements on VLCS by more than 5\%, especially with CoOp. This further validates the effectiveness of Bayes-CAL in handling both diversity shift and correlation shift.

\subsection{Ablation Studies}

\subsubsection{The Effectiveness of the Proposed Components}
For ablation studies, we remove each regularization term from our paradigm respectively and keep other components unchanged to check each component's role for OoD generalization. 
The results are shown in Table~\ref{tab:ablationstudy}.
We can see that Bayes-CAL obtains the best result on the average accuracy of the four datasets, demonstrating synergistic benefits between the regularization terms to learn invariant alignment. 
Especially on CCD and VLCS, after removing the $\mathcal{L}_{\text {IRM}}$ and $\mathcal{L}_{\text {orth}}$, the performance will drop significantly, which means better alignment can be achieved by disentangling the image features.
Note that the performance of Bayes-CAL 
on NICO and PACS is not as significant as performance on the CCD and VLCS, we have done Wilcoxon test to demonstrate the statistical significance of our results (see Appendix F).

\begin{table}
\centering
\resizebox{\linewidth}{!}{
\begin{tabular}{ccccc} 
\hline
\multirow{2}{*}{Data} & \multicolumn{3}{c}{Removed Component}        & \multirow{2}{*}{Bayes-CAL}  \\ 
\cline{2-4}
                                             & $\mathcal{L}_{\text {E}}$ & $\mathcal{L}_{\text {IRM}}$    & $\mathcal{L}_{\text {orth}}$  &                        \\ 
\hline
NICO                                         & 96.58   & 98.00  & 98.25        & 98.33        \\
CCD                                          & 68.67   & 61.33& 64.33          & 69.00           \\
PACS                                         & 91.56  & 91.77  & 91.20          & 91.82            \\
VLCS                                         & 75.43  & 73.05  & 75.90        & 78.06              \\
\hline
Average                                      & 83.06      &    81.04     &  82.42             & \textbf{84.30} \\
\hline
\end{tabular}
}
\caption{Ablation study results.}
\label{tab:ablationstudy}
\end{table}

\subsection{Base-to-New Generalization Performance}

\begin{table}
\centering
\resizebox{\linewidth}{!}{
\begin{tabular}{ccccccc} 
\hline
\multirow{2}{*}{Data}              & \multirow{2}{*}{Algorithm}   & \multicolumn{2}{c}{OoD Results}  && \multicolumn{2}{c}{I.I.D Results}
\\ 
\cline{3-4}  
\cline{6-7}
&&  Acc     & Acc*         && Acc    & Acc* \\
\hline
\multirow{4}{*}{NICO} & CoOp    & 72.10    & 88.30     && 90.55   & 96.41    \\
                      & CoCoOp  & 75.15 &88.41  && \textbf{93.64}  &  \textbf{96.93}       \\ 
                      & DPLCLIP & 46.06  &77.21   && 66.83   &86.02           \\

                      & \textbf{CAL}     & 72.12    & 89.05     &&83.29    & 95.13     \\
                      & \textbf{Bayes-CAL}   & \textbf{76.36}  & \textbf{93.26}    && 89.45 & 96.27 \\ 
\hline
\multirow{4}{*}{PACS} & CoOp    & 55.18    & 87.78    && 72.39   & 89.83      \\
                      & CoCoOp  & 58.16    & 83.87     && 59.51  & 75.97       \\ 
                      & DPLCLIP &35.96  &53.67  && 47.98  & 70.06          \\

                      & \textbf{CAL}     & 57.30     & \textbf{90.73}     && 71.50    & 91.19      \\
                      & \textbf{Bayes-CAL}  & \textbf{60.40} & 89.85 && \textbf{72.97}  & \textbf{92.54} \\
\hline
\end{tabular}
}
\caption{Base-to-new generalization performances. (Acc denotes test accuracy, and Acc* denotes the proportion of correctly identified samples with high prediction confidence. The higher the ACC*, the higher the security of the method.)}
\label{base-to-new}
\end{table}

\begin{figure*}
\centering
\includegraphics[width=0.99\textwidth]{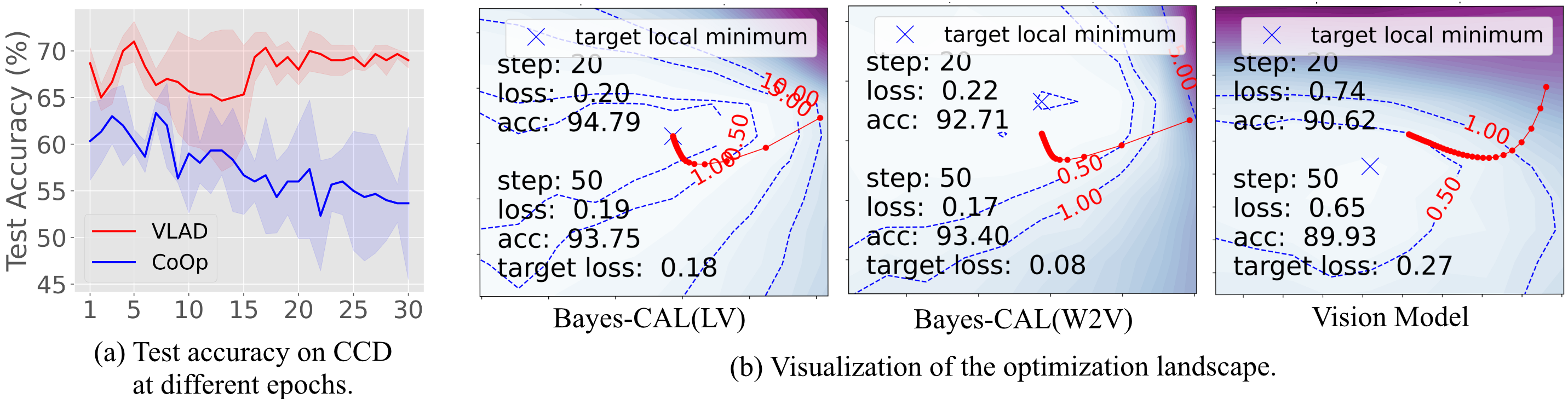}
\caption{(a) Test accuracy on CCD at different epochs. The test accuracy of Bayes-CAL is more stable than CoOp as the training epoch increases.
(b) Visualization of the optimization landscape. The x-axis and y-axis represent the first and the second principal components of the parameters, respectively. The red lines are the corresponding  optimization trajectories. Without the pre-trained text encoder, Bayes-CAL(LV) and Bayes-CAL(W2V) still see much faster convergence and lower target loss compared to the conventional visual model. }
\label{landscape}
\end{figure*}

To evaluate the effectiveness of the Bayesian method, we compare Bayes-CAL, CAL (the model that removes the Bayesian method from Bayes-CAL), CoOp, CoCoOp, and DPLCLIP in the base-to-new generalization setting on NICO and PACS. In this paper, we randomly split each dataset into the base and new sets. Following CoCoOp, the prompts are learned from the base classes (16 shots, CTP as "end", CSC as "False", and 4 context tokens), and the learnable contexts are initialized as "a photo of a". 
Under a 20-times random search
for each of the 3 different data trials for additional hyper-parameters of Bayes-CAL and CAL, the base-to-new generalization performances are evaluated on in-distribution (I.I.D) new classes and out-of-distribution (OoD) new classes. 
We also utilize the thresholding-based method to detect examples with low prediction confidence, where the threshold is selected by 95\% correctly classified validation examples are detected into examples with high prediction confidence. We remove predictions that were misclassified with low confidence, i.e., those with probabilities lower than the selected threshold, and then recalculate the test accuracy (denoted as Acc*).
The results can be seen in Table~\ref{base-to-new}. It is shown that Bayes-CAL significantly outperforms CAL on the test accuracy of new classes, especially for NICO. The results strongly justify the generalizability of the Bayesian method. Moreover, Bayes-CAL yields more stable 
I.I.D and OoD base-to-new generalization 
results than CoCoOp, which is dedicated to improving generalization to unseen classes. More details are in Appendix D.

\begin{table}
\centering
\resizebox{\linewidth}{!}{
\begin{tabular}{cccccc} 
\hline
Indicator         & CoOp       & PL  & LV  & W2V  & Vision  \\ 
\hline
Final Loss     & 0.24  & 0.28                  & \textbf{0.18}      & 0.21    & 0.74   \\
Target Loss     & 0.23  & 0.26                  & 0.15      & \textbf{0.08}    & 0.29   \\
Test Acc          & 95.58   & \textbf{96.08}          & 91.42         & 89.33   & 92.92  \\
\hline
\end{tabular}
}
\caption{Results of few-shot learning on NICO's subclasses. Bayes-CAL instantiated by prompt learning, learnable vectors, and Word2Vector is abbreviated as PL, LV, and W2V, respectively. Final Loss is the training loss at epoch 50.}
\label{different_models}
\end{table}




\subsection{Insight into the Superiority of Alignment Learning}

We have illustrated the working mechanism of alignment learning on few-shot OoD data in Figure~\ref{pipeline} (c). In this section, we elaborate on it by analyzing its convergence speed to gain insight into the superiority of alignment learning.

We still incorporate the image encoder of CLIP into the image feature extractor, while substituting the text branches with two methods: directly utilizing learnable vectors (LV) and Word2Vector with a two-layer multi-layer perception (W2V). We compare them with the Bayes-CAL instantiated by prompt learning (Bayes-CAL(PL)) and the conventional visual paradigm. Based on the same pre-trained model of CLIP's image encoder, the task-specific layers in the conventional visual model are instantiated by a three-layer multi-layer perception.

The subclasses in NICO are mostly unseen in the pre-training data of CLIP. Setting the hyper-parameters of Bayes-CAL as (0.1, 0, 0.1) and the max epoch as 50, 
we conduct a 19-way 16-shot training on NICO with 19 subclasses and 4 environments from training domains.
We report the corresponding training performance (final training loss at epoch 50) and the test accuracy in Table~\ref{different_models}. It is shown that: 1) Instantiated by LV and W2V, Bayes-CAL achieves lower final training loss. It indicates that the two instances of Bayes-CAL can still achieve fast convergence of the alignment learning without the pre-trained text encoder. 2) A wider minimum (lower target loss) can be achieved after reshaping the semantic space by LV and W2V, which also demonstrates the generality of our framework.

\subsubsection{Visualization of the Optimization Landscape}
We show the optimization trajectories and loss landscapes for Bayes-CAL(LV) and the conventional visual model in Figure~\ref{landscape}. It is observed that without the pre-trained text encoder, the instances of Bayes-CAL are close to the local minimum at epoch 50, while the conventional visual model is far from the target local minimum. Note that, Bayes-CAL(LV) has much fewer parameters (0.04 million) than the conventional model (0.28 million) to optimize. So we can say the fast learning ability is essentially caused by alignment learning.

\section{Conclusion and Discussion}
In this paper, we have proposed the Bayes-CAL method with invariant risk minimization and gradient orthogonalization loss under the Bayesian framework to tackle the few-shot OoD generalization dilemma. Numerical results not only show Bayes-CAL achieves robust OoD generalization performances under two-dimensional distribution shifts, but also reach more stable generalization performance on unseen classes under the Bayesian framework.
To the best of our knowledge, it is the first work that investigates few-shot OoD generalization by Bayesian alignment learning. This may serve as a foothold for future research in foundation models for few-shot OoD generalization. 

\section{Acknowledgements}
Nanyang Ye was supported in part by National Natural Science Foundation of China under Grant No.62106139, 
in part by National Key R\&D Program of China 2022YFB3904204, in part by National Natural Science Foundation of China under Grant (No.42050105, 61960206002, 62061146002, 62020106005).

\bibliography{main}

\clearpage

\appendix
\section{A. Model selection methods and the hyper-parameter search spaces}
Among the three model selection methods we used, training-domain validation, test-domain validation, and OoD validation are concisely described as follows:
\begin{itemize}
    \item \textbf{Training-domain validation} This strategy assumes that the training and test examples follow a similar distribution. We train models using the training subsets, and choose the model maximizing the accuracy on the union of validation subsets. 
    \item \textbf{Test-domain validation} We choose the model maximizing the accuracy on a validation set that follows the distribution of the test domain. We allow one query (the last checkpoint) per choice of hyper-parameters, disallowing early stopping.
    \item \textbf{OoD validation} This strategy assumes that the models generalizing well on the OoD validation set also generalize well on the test set. We choose the model maximizing the accuracy on a validation set that follows neither the distribution of the training domain nor the test domain. 
\end{itemize}

In addition to the four datasets involved in the paper, we also conduct experiments on ColoredMNIST \cite{arjovsky2019invariant} and OfficeHome \cite{venkateswara2017deep} to demonstrate the statistical significance of the results obtained by the Bayes-CAL.
For ColoredMNIST, we employ test-domain validation. For OfficeHome, we use training-domain validation.

The hyperparameter search spaces are presented in Table \ref{hp}. With 16 shots and 16 context tokens, the search process for CLIP-based competitors involves at least 4 times experiments for each of 3 pairs of weight initialization and training-validation data. For CoOp and CoCoOp, the class token position (CTP) is set to ``end'' or ``middle'' and the class-specific context (CSC) is ``True'' or ``False''. Unless otherwise specified, hyperparameters for DPLCLIP\cite{APCLIP} are set as their original default settings. Moreover, for all base-to-new generalization experiments, CSC is ``False'' and CTP is ``end'' as in CoCoOp\cite{zhou2022cocoop}.

\begin{table}
\centering
\caption{Hyper-parameters and distributions for random search.}
\resizebox{\linewidth}{!}{
\begin{tabular}{llll} 
\hline
Method & Dataset & Hyperparameter & Random distribution \\ 
\hline
\multirow{27}{*}{Bayes-CAL} & \multirow{4}{*}{PACS} & $\lambda_1$ & $10^{\text{Uniform}(-4, -1)}$ \\
 & & $\lambda_2$ & $10^{\text{Uniform}(-1, 0)}$ \\
 & & $\lambda_3$ & $10^{\text{Uniform}(-4, -1)}$ \\
 & & CTP & end \\
\cline{2-4}
 & \multirow{4}{*}{OfficeHome} & $\lambda_1$ & $10^{\text{Uniform}(-3, 0)}$ \\
 & & $\lambda_2$ & $10^{\text{Uniform}(-2, 0)}$ \\
 & & $\lambda_3$ & $10^{\text{Uniform}(-3, 0)}$ \\
 & & CTP & end \\                   
\cline{2-4}
 & \multirow{4}{*}{VLCS} & $\lambda_1$ & $10^{\text{Uniform}(-2, -1)}$ \\
 & & $\lambda_2$ & $10^{\text{Uniform}(-1, 0)}$ \\
 & & $\lambda_3$ & $10^{\text{Uniform}(-2, -1)}$ \\
 & & CTP & end \\ 
\cline{2-4}
 & \multirow{6}{*}{ColoredMNIST} & $\lambda_1$ & $10^{\text{Uniform}(-3, 0)}$ \\
 & & $\lambda_2$ & $10^{\text{Uniform}(-1, 1)}$ \\
 & & $\lambda_3$ & $10^{\text{Uniform}(-3, 0)}$ \\
 & & CTP & Random Choice(end, middle) \\
 & & CSC & Random Choice(True, False) \\ 
\cline{2-4}
 & \multirow{4}{*}{NICO} & $\lambda_1$ & $10^{\text{Uniform}(-3, 0)}$ \\
 & & $\lambda_2$ & $10^{\text{Uniform}(-2, 0)}$ \\
 & & $\lambda_3$ & $10^{\text{Uniform}(-3, 0)}$ \\
 & & CTP & end\\
\cline{2-4}
 & \multirow{4}{*}{CCD} & $\lambda_1$ & $10^{\text{Uniform}(-1, 0)}$ \\
 & & $\lambda_2$ & $10^{\text{Uniform}(-1, 1)}$ \\
 & & $\lambda_3$ & $10^{\text{Uniform}(-1, 0)}$ \\
 & & CTP & Random Choice(end, middle) \\ 
\hline
\multirow{2}{*}{CoOp and CoCoOp} & \multirow{2}{*}{All datasets} & CSC & Random Choice(True, False) \\
 & & CTP & Random Choice(end, middle) \\
\hline
\multirow{5}{*}{DPLCLIP} & \multirow{5}{*}{All datasets} & CSC & False \\
 & & CTP & end \\
 & & mlp\_depth & Random Choice(2, 3) \\
 & & mlp\_drop & $\text{Uniform}(0, 0.1)$ \\
 & & mlp\_width & 512 \\
\hline                                 
\end{tabular}
}
\label{hp}
\end{table}

\section{B. Framework of the Bayes-CAL and the task-specific networks instantiated by various methods}

\begin{figure}[!t]
\begin{center}
\centerline{\includegraphics[width=0.5\textwidth]{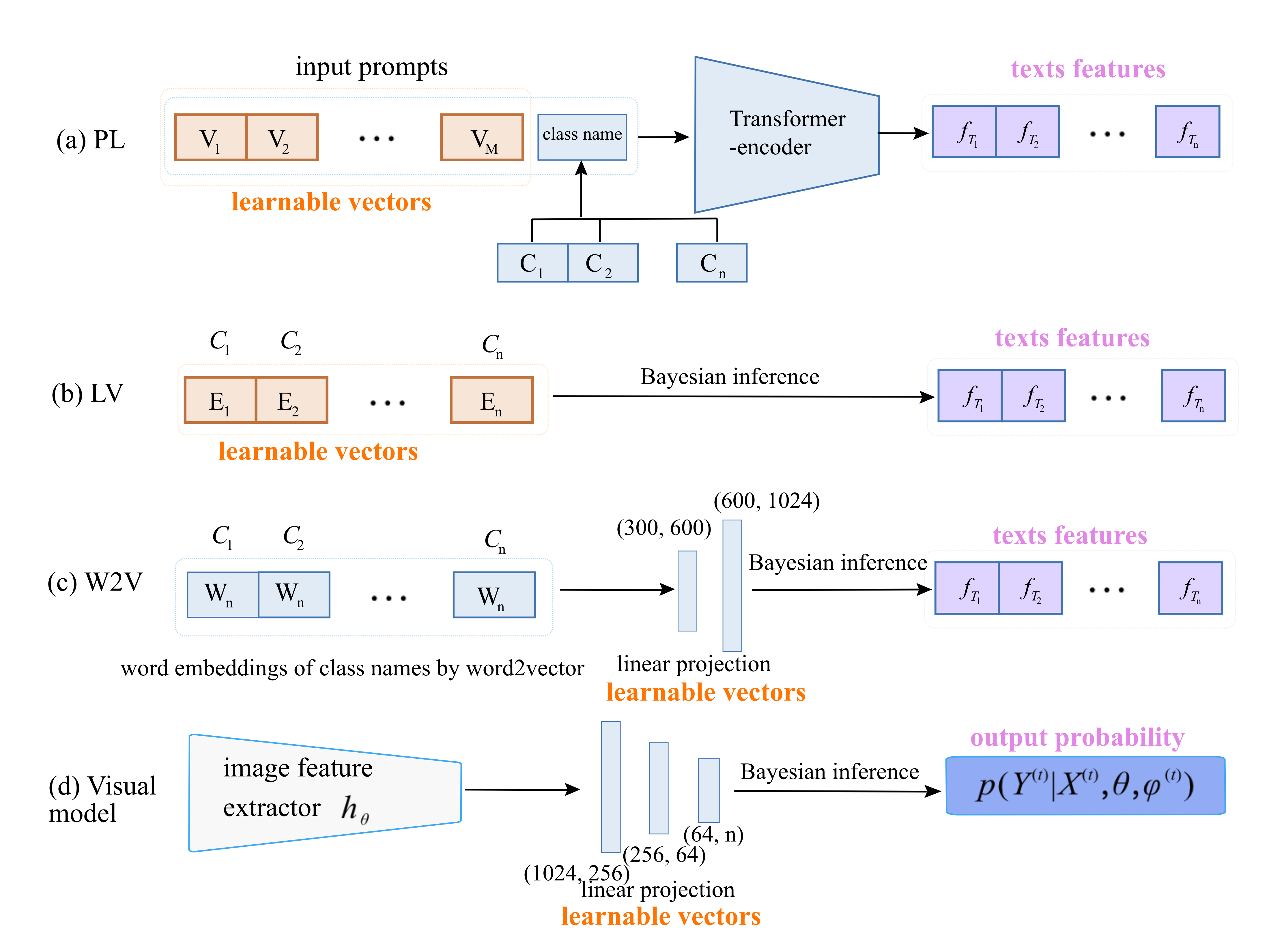}}
\caption{The three instances of the task-specific network in the Bayes-CAL and the task-specific network in the visual model.}
\label{text-branches}
\end{center}
\end{figure}

The general framework of the Bayes-CAL can be seen in Algorithm~\ref{alg:Framwork}. Combined with Figure~\ref{text-branches} (a)-(c), we elaborate the three instances of the task-specific network--Prompt Learning (PL), Learnable Vectors (LV), and word2vector (W2V) as follows:
\begin{itemize}
    \item \textbf{Prompt Learning (PL)} {As shown in Figure~\ref{text-branches} (a), the task-specific network $g_T(Y, \varphi)$ generates input prompts to the text encoder, which is designed as $g_T(Y, \varphi)=[\mathrm{V}]_{1}[\mathrm{V}]_{2} \ldots[\mathrm{V}]_{M}[\text { class name }]$. Each learnable vector $[V]_m (m \in {1,\cdots, M})$ is a vector with the same dimension $d$, $M$ is a hyper-parameter specifying the number of context tokens, and the class token position (end or middle) is the position of $[\text { class name }]$ in the prompts. In this paper,  $\{[\mathrm{V}]_{1}, \dots, [\mathrm{V}]_{M}\}$ are learnable $M\times512$-dimensional vectors, denoted as the task-specific parameter $\varphi$. }
    
    \item \textbf{Learnable Vectors (LV)} {As shown in Figure~\ref{text-branches} (b), this method substitutes the whole text branches with learnable $n\times 1024$-dimensional raw vector $\mathrm{E}_{i} (i=1, \dots, n)$ ($n$ is the number of class names) with random initialization. The raw vector $\mathrm{E}_{i}$ is corresponded to the class name $\mathrm{C}_{i} (i=1, \dots, n)$, respectively. 
    Under Bayesian variation inference, the variational distribution of the word embeddings is estimated, and finally, the texts features $\boldsymbol{f}_T$ are sampled from the variational distribution to achieve alignment with the image features.}
    
    \item \textbf{word2vector (W2V)} {As shown in Figure~\ref{text-branches} (c), this method is based on the 300-dimensional word embeddings from word2vector learned on billions of words from Wikipedia unannotated texts. We input these word embeddings to a two-layer multi-layer perception and project them into 1024-dimensional vectors. Similarly, we also estimate the variational distribution of these output 1024-dimensional vectors and then sample the texts features form  the variational distribution to achieve alignment with the image features.}
    
\end{itemize}

The conventional visual network is depicted in Figure~\ref{alg:Framwork} (d). The feature extractor $h_\theta$ and the shared parameter $\theta$ are fixed across downstream tasks, incorporating the same pre-trained image encoder of CLIP. We simply instantiate its task-specific network as a three-layer multi-layer perception as often adopted in deep learning research.

\begin{algorithm}
\caption{Framework of the proposed Bayes-CAL.}
\label{alg:Framwork}
\textbf{Input}: Training data $D_{train}$; validation data $D_{val}$; test data $D_{test}$; environment names $E$
and category names $C$; maximum epoch $M_{max}$, model selection method $V$.\\
\textbf{Output}: Task-specific parameters in category branch $\varphi_C$; task-specific parameters in environment branch $\varphi_E$ and $\varphi_E$; OoD test performance.
\begin{algorithmic}[1] 
\WHILE{$t \in \{1, \dots, M_{max} \}$}
\STATE Generate task-specific inputs by $g_{T_C}(Y_C, \varphi_C)$ based on $\varphi_C$ that is sampled from its variation distribution $q_{\varphi_C}$; Similarly, generate task-specific inputs by $g_{T_E}(Y_E, \varphi_E)$;
\STATE $f_{T_C} = f_T(X, \theta_T, g_{T_C}(Y_C, \varphi_C))$\\
$f_{T_E} = f_T(X, \theta_T, g_{T_E}(Y_E, \varphi_E))$
\STATE $f_{I} = h_I(X, \theta_I)$;
\STATE Compute alignment scores for every category name and environment name: $S(\left<f_{T_C}, f_I\right>)$ and $S(\left<f_{T_E}, f_I\right>)$;
\STATE Compute the training loss defined in Equation~(\ref{objective}) and update $\varphi_C$ and $\varphi_E$;
\ENDWHILE
\STATE Based on the validation performance, select the best model by the model selection method $V$;
\STATE Evaluate the OoD generalization performance based on the best model.
\STATE \textbf{return} $q_{\varphi_C}$, $q_{\varphi_E}$, and test accuracy.
\end{algorithmic}
\end{algorithm}

\section{C. Visualization of the visual spaces}

\begin{figure}
\begin{center}
\centerline{\includegraphics[width=0.5\textwidth]{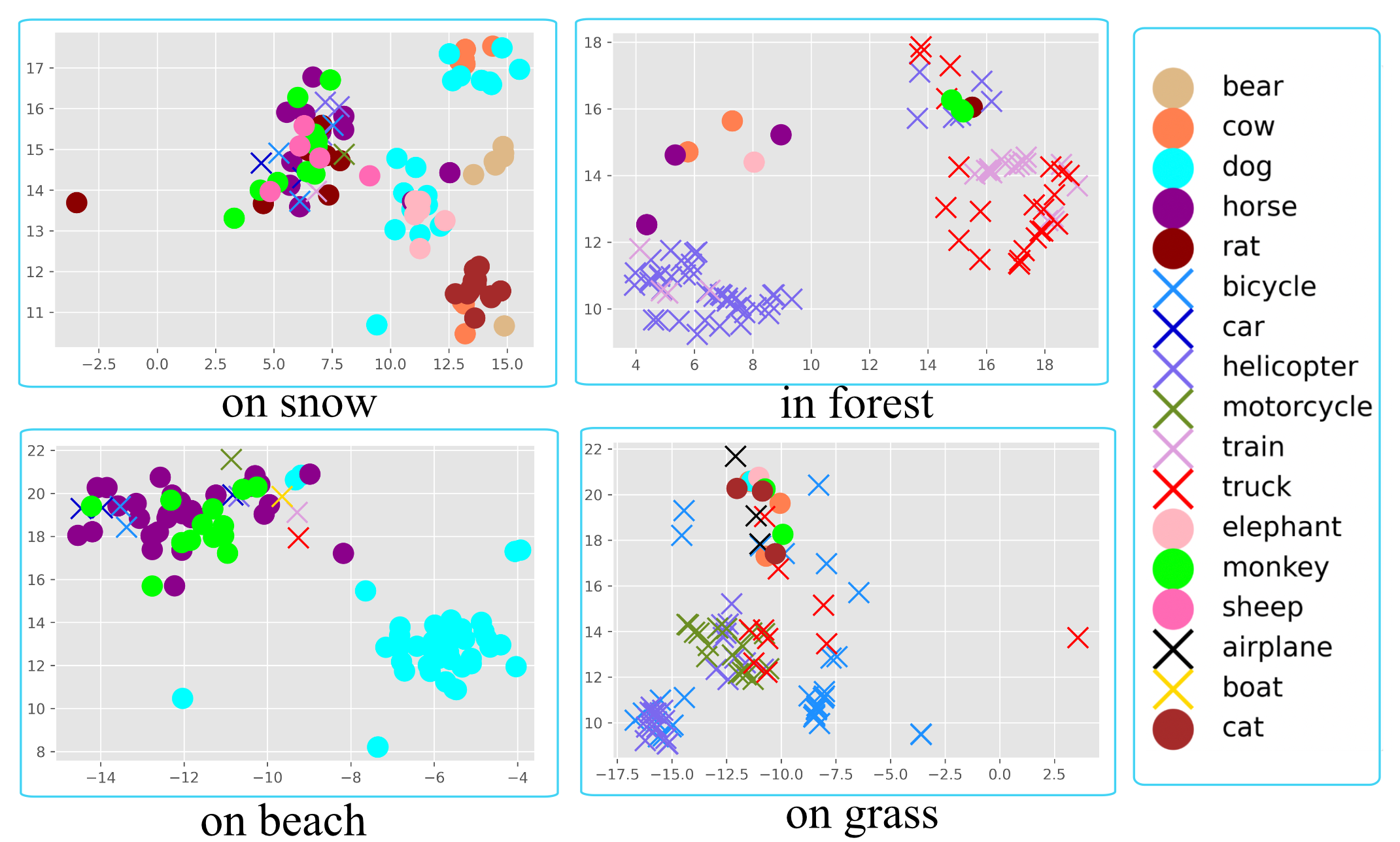}}
\caption{Visualization of the image features by 2D t-SNE. 
Categories belonging to animals and vehicles are represented by dots and crosses in different colors, respectively. 
}
\label{TSNE1}
\end{center}
\end{figure}

We illustrate the 2D t-SNE embedding of image features $\boldsymbol{f_I}$ on NICO's test sets in Figure~\ref{TSNE1}.
Consistent with the analysis above, visually similar classes are embedded closer together, while image features belonging to different super-categories overlap in each environment.

\section{D. Datasets and metrics for base-to-new generalization}
We give detailed information about data splits and evaluation metrics of base-to-new generalization.

\subsection{Data splits of NICO and PACS}
We randomly split the classes into base classes and new classes for each dataset. The class split results can be seen in Table~\ref{split}. 

\begin{table*}
\centering
\caption{Class split information of NICO and PACS}
\begin{tabular}{lll} 
\hline
Dataset & Base                                                                                                                                    & New                                       \\ 
\hline
PACS    & dog, elephant, giraffe, guitar, horse                                                                                                   & house, person                             \\
\hline
NICO    & \begin{tabular}[c]{@{}l@{}}bear, cow, dog, horse, rat, bicycle, bus, car, \\helicopter, motorcycle, train, truck, elephant\end{tabular} & monkey, sheep, airplane, bird, boat, cat  \\
\hline
\end{tabular}
\label{split}
\end{table*}

\subsection{Metrics for base-to-new generalization}
\paragraph{I.I.D\_Acc} I.I.D\_Acc is the text accuracy on the novel classes that come from the same domains as the training data.
\paragraph{OoD\_Acc} 
OoD\_Acc is the text accuracy on the novel classes that come from the text domain.
\paragraph{I.I.D\_Acc* and OoD\_Acc*} Confidence in neural network predictions became increasingly important. Basic neural networks typically give a normalized prediction vector without certainty estimates or suffer from over or under confidence. When it comes to OoD samples, out-of-domain uncertainty will arise, representing the uncertainty related to inputs drawn from new domains. 
In practice, detecting whether an example is recognized with high confidence is crucial to enable a system to reject such samples or alert users. It is essential to quantify the model's prediction uncertainty or detect unconfident predictions, especially in the case that the text classes have not been observed in training data.

In this paper, we use the output probability (the maximum element in the normalized prediction vector) as the prediction confidence score and utilize the thresholding-based method to detect examples with low prediction confidence.
The threshold is selected by 95\% correctly classified validation examples are detected into examples with high prediction confidence.
That is, the threshold is the 0.95 quantile of output probabilities of these correctly classified validation examples.
Based on the threshold, we refuse to classify the samples with output probabilities below the threshold and recompute the test accuracy (Acc*), which denotes the proportion of correctly identified samples in all predictions with confidence scores higher than the threshold. 
And hence, we obtain the I.I.D\_Acc* and OoD\_Acc* for I.I.D\_Acc and OoD\_Acc, respectively. The higher the Acc*, the higher the security of the classification results.

According to Table 4 in the paper, it is observed that Bayes-CAL
achieved the best average results, which demonstrates the effectiveness of the Bayes-CAL for better few-shot OoD generalize on the unseen classes under both diversity shift and correlation shift. Especially on NICO, the significant
improvements of Bayes-CAL compared to CAL provide solid proof that the variational approximation of the prompts can embed more information compared to the deterministic estimation, and thus avoid overfitting on base classes and improves the base-to-new generalization ability.



\begin{table}[!t]
\centering
\caption{Few-shot OoD generalization performances on ColoredMNIST and OfficeHome.}
\label{addition}
\resizebox{\linewidth}{!}{
\begin{tabular}{lllll} 
\hline
             & CLIP  & CoOp   & CoCoOp  & Bayes-CAL           \\ 
\hline
ColoredMNIST &47.10±0.00  & 9.50±2.51 &17.44±4.91   & \textbf{47.88±0.62}       \\
OfficeHome    &69.00±0.00  &62.39±0.69  &72.27±0.54  & \textbf{74.20±0.02}   \\
\hline
\end{tabular}
}
\end{table}

\begin{table}[!t]
\centering
\caption{Ablation study results on ColoredMNIST and OfficeHome.}
\resizebox{\linewidth}{!}{
\begin{tabular}{ccccc} 
\hline
\multirow{2}{*}{Dataset} & \multicolumn{3}{c}{Removed Component}        & \multirow{2}{*}{Bayes-CAL}  \\ 
\cline{2-4}
                                             & $\mathcal{L}_{\text {E}}$ & $\mathcal{L}_{\text {IRM}}$    & $\mathcal{L}_{\text {orth}}$  &                        \\ 
\hline
ColoredMNIST                                         & 42.56±6.41  & 6.71±2.39        & 34.88±10.89        & \textbf{47.88±0.62}        \\
OfficeHome                                         & 73.75±0.12   & 73.72±0.21  & 73.68±0.25          & \textbf{74.20±0.02}              \\
\hline
\label{add_abla}
\end{tabular}
}
\end{table}

\begin{table}[!t]
\centering
\caption{The Wilcoxon test results for pairwise comparisons.}
\resizebox{\linewidth}{!}{
\begin{threeparttable}
\begin{tabular}{cccc} 
\hline
\multicolumn{2}{l}{Comparison to CLIP-based models} & \multicolumn{2}{l}{Comparison to models in the ablation study}  \\ 
\hline
CLIP vs Ours    & 0.03                               & $-\mathcal{L}_{\text {E}}$ vs Ours & 0.04                \\
CoOp vs Ours    & 0.03                               & $-\mathcal{L}_{\text {IRM}}$ vs Ours  & 0.03                \\
CoCoOp vs Ours  & 0.03                               & $-\mathcal{L}_{\text {orth}}$ vs Ours      & 0.04                                   \\
\hline
\end{tabular}
\begin{tablenotes} 
        \footnotesize 
        \item $-\mathcal{L}_{\cdot}$ means removing the corresponding loss.
      \end{tablenotes} 
    \end{threeparttable}
\label{wilcoxon}
}
\end{table}

\section{E. Comparison with other CLIP-based models based on image feature adapter}
In addition to fine-tuning language features, an alternative path is to incorporate feature adapters into the visual branch. For example, CLIP-Adapter \cite{gao2021clip} appends a small number of additional learnable bottleneck linear layers to CLIP’s language and visual branches while keeping the original CLIP backbone frozen during few-shot fine-tuning. A training-free adaption method Tip-Adapter \cite{zhang2021tip} constructs the visual feature adapter via a key-value cache model from the few-shot training set, and updates the prior knowledge encoded in CLIP by feature retrieval. Moreover, the performance
of Tip-Adapter can be further boosted by fine-tuning
such properly initialized adapter for only a few epochs
with super-fast convergence speed.

\begin{table}[!t]
\centering
\caption{Performances of few-shot OoD generalization of Tip-Adapter-F and Bayes-CAL.}
\resizebox{\linewidth}{!}{
\begin{tabular}{llllll} 
\hline
             & NICO  & CCD   & PACS  & VLCS  & Average         \\ 
\hline
Tip-Adapter-F & 97.91 & 62.33 & 92.12 & 77.22 & 82.40           \\
Bayes-CAL    & 98.33 & 69.00 & 91.82 & 78.06 & \textbf{84.30}  \\
\hline
\end{tabular}
}
\label{Tip}
\end{table}

We evaluate Tip-Adapter's few-shot OoD generalization performance on the four OoD datasets--NICO, CCD, PACS, and VLCS. Under the same experiment protocols, we set the search scale and the search step of its hyper-parameter $\alpha$ and $\beta$ as (20, 20) and (200, 20), respectively. The initial value of $\alpha$ and $\beta$ is set as 1 and 5, respectively.
The average text accuracies of the 3 random experiments can be seen in Table~\ref{Tip}. We can see that our Bayes-CAL achieves better average results than Tip-Adapter by around 2\%.

\section{F. The Wilcoxon test results for pairwise comparisons}
We conduct experiments on ColoredMNIST and OfficeHome, which have been introduced in Appendix A.
The corresponding results can be seen in Table~\ref{addition} and Table~\ref{add_abla}. {The much better performance of the Bayes-CAL on ColoredMNIST further provides solid proof that Bayes-CAL can achieve better alignment based on the disentangled image features.}
We do the Wilcoxon test based on the results on all six datasets, and the results can be seen in Table~\ref{wilcoxon}.
It is shown that the equivalence hypotheses are all rejected with very low p-values, which validates the statistical significance of Bayes-CAL's better results compared to CLIP-based models and the methods in the ablation study.

\end{document}


\appendix
\section{A. Model selection methods and the hyper-parameter search spaces}
Among the three model selection methods we used, training-domain validation, test-domain validation, and OoD validation are concisely described as follows:
\begin{itemize}
    \item \textbf{Training-domain validation} This strategy assumes that the training and test examples follow a similar distribution. We train models using the training subsets, and choose the model maximizing the accuracy on the union of validation subsets. 
    \item \textbf{Test-domain validation} We choose the model maximizing the accuracy on a validation set that follows the distribution of the test domain. We allow one query (the last checkpoint) per choice of hyper-parameters, disallowing early stopping.
    \item \textbf{OoD validation} This strategy assumes that the models generalizing well on the OoD validation set also generalize well on the test set. We choose the model maximizing the accuracy on a validation set that follows neither the distribution of the training domain nor the test domain. 
\end{itemize}

In addition to the four datasets involved in the paper, we also conduct experiments on ColoredMNIST \cite{arjovsky2019invariant} and OfficeHome \cite{venkateswara2017deep} to demonstrate the statistical significance of the results obtained by the Bayes-CAL.
For ColoredMNIST, we employ test-domain validation. For OfficeHome, we use training-domain validation.

The hyperparameter search spaces are presented in Table \ref{hp}. With 16 shots and 16 context tokens, the search process for CLIP-based competitors involves at least 4 times experiments for each of 3 pairs of weight initialization and training-validation data. For CoOp and CoCoOp, the class token position (CTP) is set to ``end'' or ``middle'' and the class-specific context (CSC) is ``True'' or ``False''. Unless otherwise specified, hyperparameters for DPLCLIP\cite{APCLIP} are set as their original default settings. Moreover, for all base-to-new generalization experiments, CSC is ``False'' and CTP is ``end'' as in CoCoOp\cite{zhou2022cocoop}.

\begin{table}
\centering
\caption{Hyper-parameters and distributions for random search.}
\resizebox{\linewidth}{!}{
\begin{threeparttable}
\begin{tabular}{llll} 
\hline
Method & Dataset                       & Hyperparameter                                    & Random distribution                   \\ 
\hline
\multirow{27}{*}{Bayes-CAL}      & \multirow{4}{*}{PACS}         & \vcell{$\lambda_1$}                                & \vcell{$10^{\mbox{Uniform}(-4, -1)}$}  \\
                                 &                               & $\lambda_2$                                        & $10^{\mbox{Uniform}(-1, 0)}$       \\
                                 &                               & $\lambda_3$                                        & $10^{\mbox{Uniform}(-4, -1)}$   \\
                                 &                               & CTP
                                 & end \\
                                  
\cline{2-4}
                                 & \multirow{4}{*}{OfficeHome}   & $\lambda_1$                                        & $10^{\mbox{Uniform}(-3, 0)}$          \\
                                 &                               & $\lambda_2$                                        & $10^{\mbox{Uniform}(-2, 0)}$          \\
                                 &                               & $\lambda_3$                                         & $10^{\mbox{Uniform}(-3, 0)}$          \\
                                 &                               & CTP
                                 & end \\                   
\cline{2-4}
                                 & \multirow{4}{*}{VLCS}         & $\lambda_1$                                        & $10^{\mbox{Uniform}(-2, -1)}$          \\
                                 &                               & $\lambda_2$                                        & $10^{\mbox{Uniform}(-1, 0)}$          \\
                                 &                               & $\lambda_3$                                        & $10^{\mbox{Uniform}(-2, -1)}$          \\
                                 &                               & CTP                                               & end            \\ 
\cline{2-4}
                                 & \multirow{6}{*}{ColoredMNIST} & $\lambda_1$                                        & $10^{\mbox{Uniform}(-3, 0)}$          \\
                                 &                               & $\lambda_2$                                        & $10^{\mbox{Uniform}(-1, 1)}$          \\
                                 &                               & $\lambda_3$                                        & $10^{\mbox{Uniform}(-3, 0)}$          \\
                                 &                               & CTP                                               & Random Choice(end, middle)            \\
                                 
                                 &                               & CSC                                               & Random Choice(True, False)            \\ 
\cline{2-4}
                                 & \multirow{4}{*}{NICO}         & $\lambda_1$                                        & $10^{\mbox{Uniform}(-3, 0)}$          \\
                                 &                               & $\lambda_2$                                        & $10^{\mbox{Uniform}(-2, 0)}$          \\
                                 &                               & $\lambda_3$                                        & $10^{\mbox{Uniform}(-3, 0)}$          \\
                                 &                               & CTP
                                 & end\\
                                 
\cline{2-4}
                                 & \multirow{4}{*}{CCD}          & $\lambda_1$                                        & $10^{\mbox{Uniform}(-1, 0)}$          \\
                                 &                               & $\lambda_2$                                        & $10^{\mbox{Uniform}(-1, 1)}$          \\
                                 &                               & $\lambda_3$                                        & $10^{\mbox{Uniform}(-1, 0)}$          \\
                                 &                               & CTP                                               &  Random Choice(end, middle)            \\ 
                                 
\hline
\multirow{2}{*}{CoOp and CoCoOp}      & \multirow{2}{*}{All datasets}         
                                 &  CSC                                        & Random Choice(True, False)   \\
                                 &                               & CTP
                                 & Random Choice(end, middle) \\
\cline{2-4}
\multirow{5}{*}{DPLCLIP}      & \multirow{5}{*}{All datasets}         
                                 &  CSC                                        &  False   \\
                                 &                               & CTP
                                 & end\\
                                 &          
                                 &  mlp\_depth                                        & Random Choice(2, 3)   \\
                                 &                               & mlp\_drop
                                 & ${\mbox{Uniform}(0, 0.1)}$ \\
                                 &                               & mlp\_width
                                 & 512 \\
\hline                                 
\end{tabular}
\begin{tablenotes} 
        \footnotesize 
        \item CSC of the Bayes-CAL is set as ``False'' unless otherwise specified.
      \end{tablenotes} 
    \end{threeparttable}
\label{hp}
}
\end{table}

\section{B. Framework of the Bayes-CAL and the task-specific networks instantiated by various methods}

\begin{figure}[!t]
\begin{center}
\centerline{\includegraphics[width=0.5\textwidth]{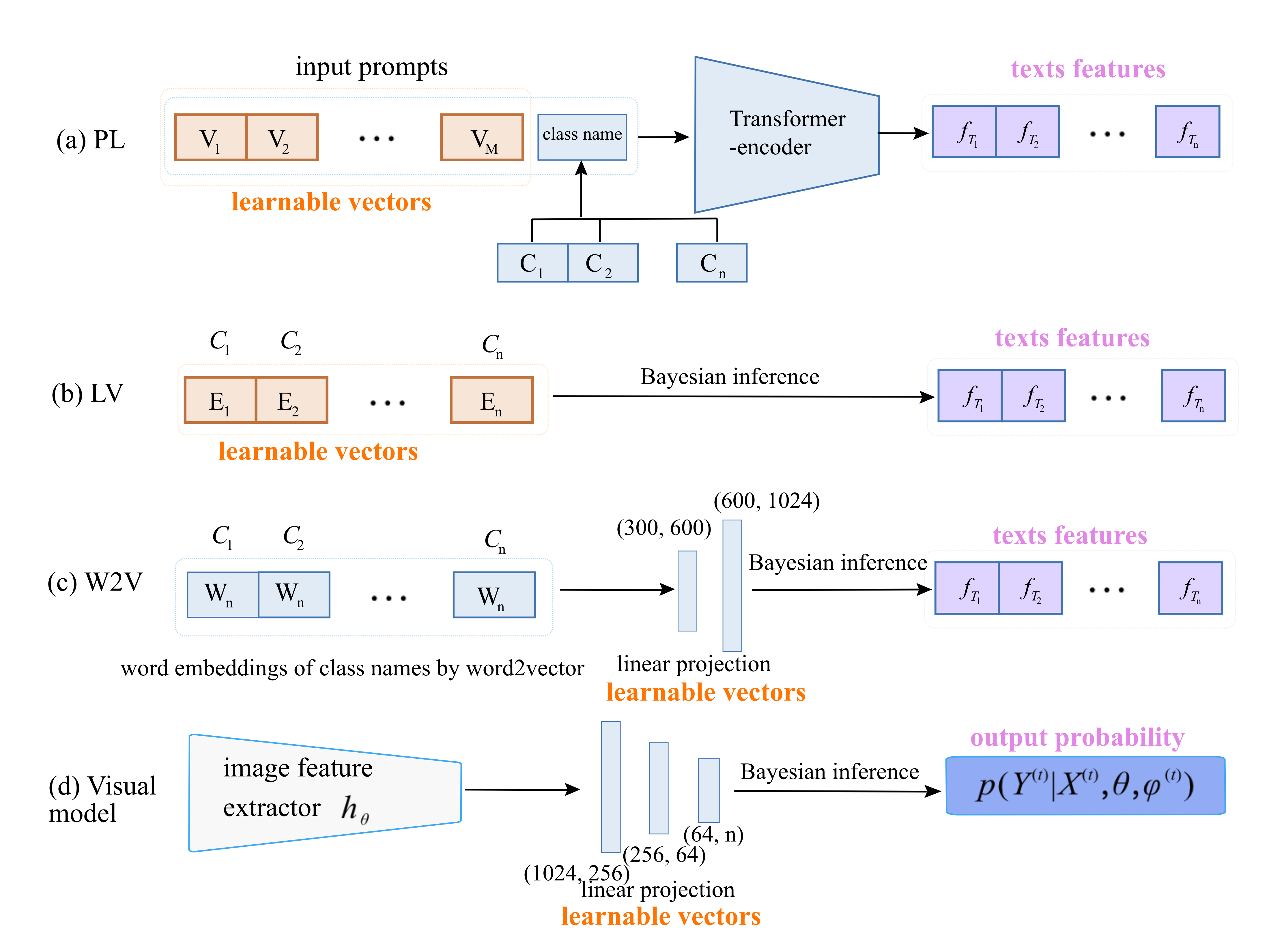}}
\caption{The three instances of the task-specific network in the Bayes-CAL and the task-specific network in visual model.}
\label{text-branches}
\end{center}
\end{figure}

The general framework of the Bayes-CAL can be seen in Algorithm~\ref{alg:Framwork}. Combined with Figure~\ref{text-branches} (a)-(c), we elaborate the three instances of the task-specific network--Prompt Learning (PL), Learnable Vectors (LV), and word2vector (W2V) as follows:
\begin{itemize}
    \item \textbf{Prompt Learning (PL)} {As shown in Figure~\ref{text-branches} (a), the task-specific network $g_T(Y, \varphi)$ generates input prompts to the text encoder, which is designed as $g_T(Y, \varphi)=[\mathrm{V}]_{1}[\mathrm{V}]_{2} \ldots[\mathrm{V}]_{M}[\text { class name }]$. Each learnable vector $[V]_m (m \in {1,\cdots, M})$ is a vector with the same dimension $d$, $M$ is a hyper-parameter specifying the number of context tokens, and the class token position (end or middle) is the position of $[\text { class name }]$ in the prompts. In this paper,  $\{[\mathrm{V}]_{1}, \dots, [\mathrm{V}]_{M}\}$ are learnable $M\times512$-dimensional vectors, denoted as the task-specific parameter $\varphi$. }
    
    \item \textbf{Learnable Vectors (LV)} {As shown in Figure~\ref{text-branches} (b), this method substitutes the whole text branches with learnable $n\times 1024$-dimensional raw vector $\mathrm{E}_{i} (i=1, \dots, n)$ ($n$ is the number of class names) with random initialization. The raw vector $\mathrm{E}_{i}$ is corresponded to the class name $\mathrm{C}_{i} (i=1, \dots, n)$, respectively. 
    Under Bayesian variation inference, the variational distribution of the word embeddings is estimated, and finally, the texts features $\boldsymbol{f}_T$ are sampled from the variational distribution to achieve alignment with the image features.}
    
    \item \textbf{word2vector (W2V)} {As shown in Figure~\ref{text-branches} (c), this method is based on the 300-dimensional word embeddings from word2vector learned on billions of words from Wikipedia unannotated texts. We input these word embeddings to a two-layer multi-layer perception and project them into 1024-dimensional vectors. Similarly, we also estimate the variational distribution of these output 1024-dimensional vectors and then sample the texts features form  the variational distribution to achieve alignment with the image features.}
    
\end{itemize}

The conventional visual network is depicted in Figure~\ref{alg:Framwork} (d). The feature extractor $h_\theta$ and the shared parameter $\theta$ are fixed across downstream tasks, incorporating the same pre-trained image encoder of CLIP. We simply instantiate its task-specific network as a three-layer multi-layer perception as often adopted in deep learning research.

\begin{algorithm}
\caption{Framework of the proposed Bayes-CAL.}
\label{alg:Framwork}
\textbf{Input}: Training data $D_{train}$; validation data $D_{val}$; test data $D_{test}$; environment names $E$
and category names $C$; maximum epoch $M_{max}$, model selection method $V$.\\
\textbf{Output}: Task-specific parameters in category branch $\varphi_C$; task-specific parameters in environment branch $\varphi_E$ and $\varphi_E$; OoD test performance.
\begin{algorithmic}[1] 
\WHILE{$t \in \{1, \dots, M_{max} \}$}
\STATE Generate task-specific inputs by $g_{T_C}(Y_C, \varphi_C)$ based on $\varphi_C$ that is sampled from its variation distribution $q_{\varphi_C}$; Similarly, generate task-specific inputs by $g_{T_E}(Y_E, \varphi_E)$;
\STATE $f_{T_C} = f_T(X, \theta_T, g_{T_C}(Y_C, \varphi_C))$\\
$f_{T_E} = f_T(X, \theta_T, g_{T_E}(Y_E, \varphi_E))$
\STATE $f_{I} = h_I(X, \theta_I)$;
\STATE Compute alignment scores for every category name and environment name: $S(\left<f_{T_C}, f_I\right>)$ and $S(\left<f_{T_E}, f_I\right>)$;
\STATE Compute the training loss defined in Equation~(\ref{objective}) and update $\varphi_C$ and $\varphi_E$;
\ENDWHILE
\STATE Based on the validation performance, select the best model by the model selection method $V$;
\STATE Evaluate the OoD generalization performance based on the best model.
\STATE \textbf{return} $q_{\varphi_C}$, $q_{\varphi_E}$, and test accuracy.
\end{algorithmic}
\end{algorithm}

\begin{table*}
\centering
\caption{Class split information of NICO and PACS}
\begin{tabular}{lll} 
\hline
Dataset & Base                                                                                                                                    & New                                       \\ 
\hline
PACS    & dog, elephant, giraffe, guitar, horse                                                                                                   & house, person                             \\
\hline
NICO    & \begin{tabular}[c]{@{}l@{}}bear, cow, dog, horse, rat, bicycle, bus, car, \\helicopter, motorcycle, train, truck, elephant\end{tabular} & monkey, sheep, airplane, bird, boat, cat  \\
\hline
\end{tabular}
\label{split}
\end{table*}

\section{C. Visualization of the visual spaces}

\begin{figure}
\begin{center}
\centerline{\includegraphics[width=0.5\textwidth]{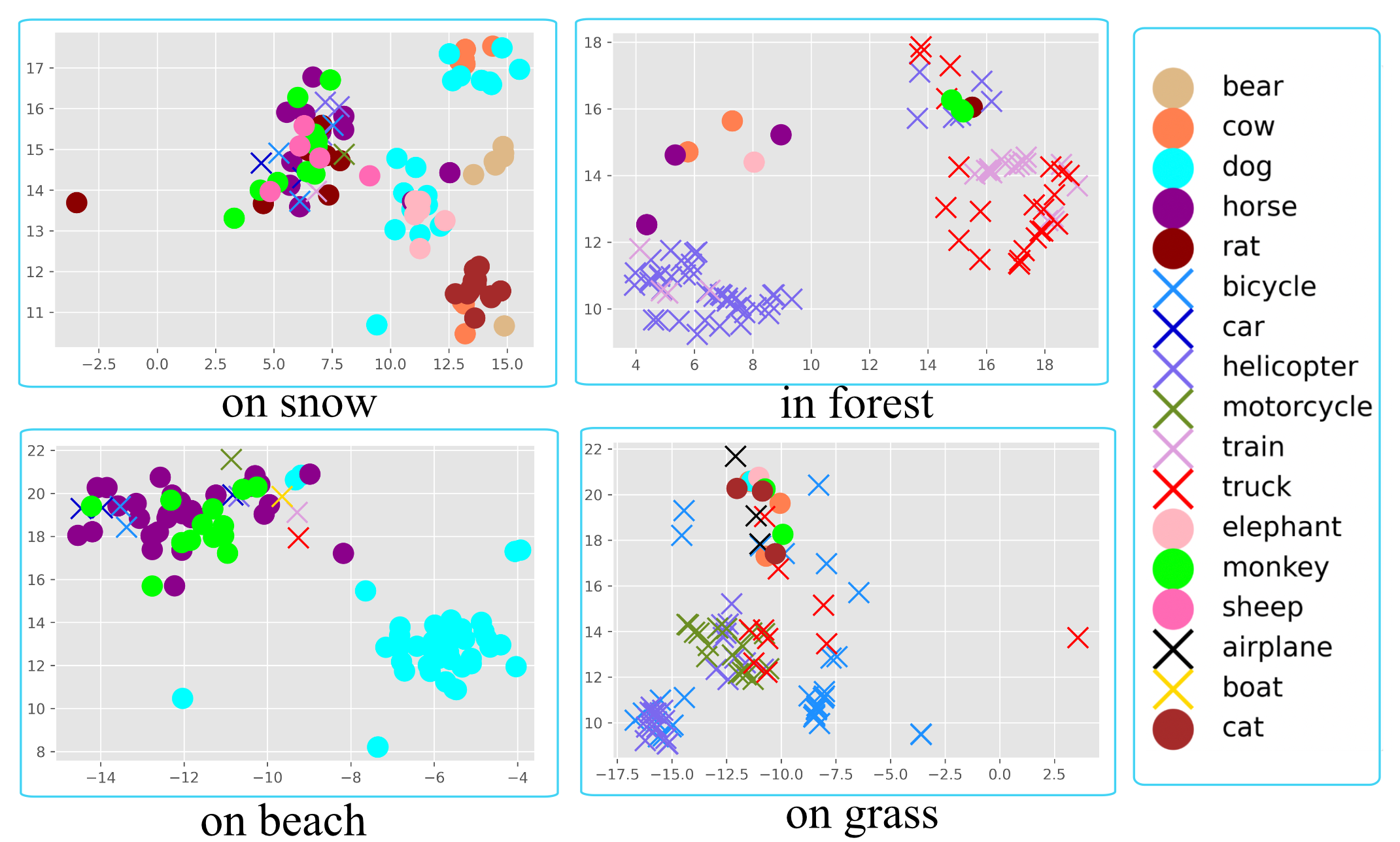}}
\caption{Visualization of the image features by 2D t-SNE. 
Categories belonging to animals and vehicles are represented by dots and crosses in different colors, respectively. 
}
\label{TSNE1}
\end{center}
\end{figure}

We illustrate the 2D t-SNE embedding of image features $\boldsymbol{f_I}$ on NICO's test sets in Figure~\ref{TSNE1}.
Consistent with the analysis above, visually similar classes are embedded closer together, while image features belonging to different super-categories overlap in each environment.

\section{D. Datasets and metrics for base-to-new generalization}
We give detailed information about data splits and evaluation metrics of base-to-new generalization.

\subsection{Data splits of NICO and PACS}
We randomly split the classes into base classes and new classes for each dataset. The class split results can be seen in Table~\ref{split}. 


\subsection{Metrics for base-to-new generalization}
\paragraph{I.I.D\_Acc} I.I.D\_Acc is the text accuracy on the novel classes that come from the same domains as the training data.
\paragraph{OoD\_Acc} 
OoD\_Acc is the text accuracy on the novel classes that come from the text domain.
\paragraph{I.I.D\_Acc* and OoD\_Acc*} Confidence in neural network predictions became increasingly important. Basic neural networks typically give a normalized prediction vector without certainty estimates or suffer from over or under confidence. When it comes to OoD samples, out-of-domain uncertainty will arise, representing the uncertainty related to inputs drawn from new domains. 
In practice, detecting whether an example is recognized with high confidence is crucial to enable a system to reject such samples or alert users. It is essential to quantify the model's prediction uncertainty or detect unconfident predictions, especially in the case that the text classes have not been observed in training data.

In this paper, we use the output probability (the maximum element in the normalized prediction vector) as the prediction confidence score and utilize the thresholding-based method to detect examples with low prediction confidence.
The threshold is selected by 95\% correctly classified validation examples are detected into examples with high prediction confidence.
That is, the threshold is the 0.95 quantile of output probabilities of these correctly classified validation examples.
Based on the threshold, we refuse to classify the samples with output probabilities below the threshold and recompute the test accuracy (Acc*), which denotes the proportion of correctly identified samples in all predictions with confidence scores higher than the threshold. 
And hence, we obtain the I.I.D\_Acc* and OoD\_Acc* for I.I.D\_Acc and OoD\_Acc, respectively. The higher the Acc*, the higher the security of the classification results.

According to Table 4 in the paper, it is observed that Bayes-CAL
achieved the best average results, which demonstrates the effectiveness of the Bayes-CAL for better few-shot OoD generalize on the unseen classes under both diversity shift and correlation shift. Especially on NICO, the significant
improvements of Bayes-CAL compared to CAL provide solid proof that the variational approximation of the prompts can embed more information compared to the deterministic estimation, and thus avoid overfitting on base classes and improves the base-to-new generalization ability.



\begin{table}[!t]
\centering
\caption{Few-shot OoD generalization performances on ColoredMNIST and OfficeHome.}
\label{addition}
\resizebox{\linewidth}{!}{
\begin{tabular}{lllll} 
\hline
             & CLIP  & CoOp   & CoCoOp  & Bayes-CAL           \\ 
\hline
ColoredMNIST &47.10±0.00  & 9.50±2.51 &17.44±4.91   & \textbf{47.88±0.62}       \\
OfficeHome    &69.00±0.00  &62.39±0.69  &72.27±0.54  & \textbf{72.74±0.47}   \\
\hline
\end{tabular}
}
\end{table}

\begin{table}[!t]
\centering
\caption{Ablation study results on ColoredMNIST and OfficeHome.}
\resizebox{\linewidth}{!}{
\begin{tabular}{ccccc} 
\hline
\multirow{2}{*}{Dataset} & \multicolumn{3}{c}{Removed Component}        & \multirow{2}{*}{Bayes-CAL}  \\ 
\cline{2-4}
                                             & $\mathcal{L}_{\text {E}}$ & $\mathcal{L}_{\text {IRM}}$    & $\mathcal{L}_{\text {orth}}$  &                        \\ 
\hline
ColoredMNIST                                         & 42.56±6.41  & 6.71±2.39        & 34.88±10.89        & \textbf{47.88±0.62}        \\
OfficeHome                                         & 72.74±0.91   & 71.70±0.50  & 72.61±0.87          & \textbf{72.74±0.47}              \\
\hline
\label{add_abla}
\end{tabular}
}
\end{table}

\begin{table}[!t]
\centering
\caption{The Wilcoxon test results for pairwise comparisons.}
\resizebox{\linewidth}{!}{
\begin{threeparttable}
\begin{tabular}{cccc} 
\hline
\multicolumn{2}{l}{Comparison to CLIP-based models} & \multicolumn{2}{l}{Comparison to models in the ablation study}  \\ 
\hline
CLIP vs Ours    & 0.03                               & $-\mathcal{L}_{\text {E}}$ vs Ours & 0.04                \\
CoOp vs Ours    & 0.03                               & $-\mathcal{L}_{\text {IRM}}$ vs Ours  & 0.03                \\
CoCoOp vs Ours  & 0.03                               & $-\mathcal{L}_{\text {orth}}$ vs Ours      & 0.04                                   \\
\hline
\end{tabular}
\begin{tablenotes} 
        \footnotesize 
        \item $-\mathcal{L}_{\cdot}$ means removing the corresponding loss.
      \end{tablenotes} 
    \end{threeparttable}
\label{wilcoxon}
}
\end{table}

\section{E. Comparison with other CLIP-based models based on image feature adapter}
In addition to fine-tuning language features, an alternative path is to incorporate feature adapters into the visual branch. For example, CLIP-Adapter \cite{gao2021clip} appends a small number of additional learnable bottleneck linear layers to CLIP’s language and visual branches while keeping the original CLIP backbone frozen during few-shot fine-tuning. A training-free adaption method Tip-Adapter \cite{zhang2021tip} constructs the visual feature adapter via a key-value cache model from the few-shot training set, and updates the prior knowledge encoded in CLIP by feature retrieval. Moreover, the performance
of Tip-Adapter can be further boosted by fine-tuning
such properly initialized adapter for only a few epochs
with super-fast convergence speed.

\begin{table}[!t]
\centering
\caption{Performances of few-shot OoD generalization of Tip-Adapter-F and Bayes-CAL.}
\resizebox{\linewidth}{!}{
\begin{tabular}{llllll} 
\hline
             & NICO  & CCD   & PACS  & VLCS  & Average         \\ 
\hline
Tip-Adapter-F & 97.91 & 62.33 & 92.12 & 77.22 & 82.40           \\
Bayes-CAL    & 98.33 & 69.00 & 91.82 & 78.06 & \textbf{84.30}  \\
\hline
\end{tabular}
}
\label{Tip}
\end{table}

We evaluate Tip-Adapter's few-shot OoD generalization performance on the four OoD datasets--NICO, CCD, PACS, and VLCS. Under the same experiment protocols, we set the search scale and the search step of its hyper-parameter $\alpha$ and $\beta$ as (20, 20) and (200, 20), respectively. The initial value of $\alpha$ and $\beta$ is set as 1 and 5, respectively.
The average text accuracies of the 3 random experiments can be seen in Table~\ref{Tip}. We can see that our Bayes-CAL achieves better average results than Tip-Adapter by around 2\%.

\section{F. The Wilcoxon test results for pairwise comparisons}
We conduct experiments on ColoredMNIST and OfficeHome, which have been introduced in Appendix A.
The corresponding results can be seen in Table~\ref{addition} and Table~\ref{add_abla}. {The much better performance of the Bayes-CAL on ColoredMNIST further provides solid proof that Bayes-CAL can achieve better alignment based on the disentangled image features.}
We do the Wilcoxon test based on the results on all six datasets, and the results can be seen in Table~\ref{wilcoxon}.
It is shown that the equivalence hypotheses are all rejected with very low p-values, which validates the statistical significance of Bayes-CAL's better results compared to CLIP-based models and the methods in the ablation study.

\bibliography{ref}